\documentclass{article}
\usepackage{spconf,amsmath,graphicx}
\usepackage{multirow}
\usepackage{bm}
\usepackage{algorithm}
\usepackage{algpseudocode}
\usepackage{amsmath}
\usepackage{graphics}
\usepackage{epsfig}
\usepackage{subfig,float}

\title{Very Deep Convolutional Neural Networks\\ for Robust Speech Recognition}
\name{Yanmin Qian$^{1,2}$, Philip C Woodland$^{2}$\thanks{This work was supported by the UK EPSRC Programme Grant EP/I031022/1 (Natural Speech Technology), the Shanghai Sailing Program No. 16YF1405300, and China NSFC projects No. 61603252.}}

\address{$^{1}$ Department of Computer Science and Engineering, Shanghai Jiao Tong University, Shanghai, China\\
			  $^{2}$Cambridge University Engineering Department, Cambridge CB2 1PZ, UK\\
{yanminqian@sjtu.edu.cn, pcw@eng.cam.ac.uk}}

\begin{document}
%
\maketitle
\begin{abstract}
This paper describes the extension and optimisation of our previous work on very deep convolutional neural networks (CNNs) for effective recognition of noisy speech in the Aurora 4 task. The appropriate number of convolutional layers, the sizes of the filters,  pooling operations and input feature maps are all modified: the filter and pooling sizes are reduced and dimensions of input feature maps are extended to allow adding more convolutional layers. Furthermore appropriate input padding and input feature map selection strategies are developed. In addition, an adaptation framework using joint training of very deep CNN with auxiliary features i-vector and fMLLR features is developed. These modifications give substantial word error rate reductions over the standard CNN used as baseline. Finally the very deep CNN is combined with an LSTM-RNN acoustic model and it is shown that state-level weighted log likelihood score combination in a joint acoustic model decoding scheme is very effective.
On the  Aurora 4 task, the very deep CNN achieves a WER of 8.81\%, further 7.99\% with auxiliary feature joint training, and 7.09\% with LSTM-RNN joint decoding.
\end{abstract}
\section{Introduction}
\label{sec:introd}

In the last few years there has been significant progress in automatic speech recognition (ASR)  due to the introduction of deep neural network (DNN) based acoustic models \cite{hinton2012deep, seide-interspeech11, dahl-ieee12}. These advancements have reduced the word error rate (WER) to a level that has allowed the successful use of ASR in many close-talking scenarios, such as voice search on smart phones and office dictation, i.e.  situations where the signal-to-noise ratio (SNR) is relatively high.  However, these systems still perform relatively poorly in noisy environments \cite{Wang12} and noise robustness is still a key issue in allowing ASR systems to have a wider range of use cases.

Several methods using the DNN framework \cite{li2014overview, shilin2014joint} have been proposed to handle the difficult problem of mismatch between training and subsequent testing in a noisy speech environment, and some improvements can be obtained. However, there is still a large performance gap when compared to the close-talking scenario with a high SNR. More recently, some alternative neural networks structures have been explored, and the Convolutional Neural Networks (CNN) is one of the most promising types. The use of CNNs has been shown to yield lower WERs than standard fully connected feed-forward DNNs for several tasks \cite{Sainath13,sainath2015convolutional}.


The majority of previous work using CNNs for speech recognition have only used up to two convolutional layers, and while \cite{Sainath13} attempted to  use three convolutional layers, this gave poorer  performance. Based on the previous work using CNNs for ASR, the configuration described in \cite{Sainath15a} is usually used as the standard CNN architecture for speech recognition, with two convolutional layers followed by four fully connected layers. Recently, the computer vision community \cite{Szegedy14,Simonyan14,he2015delving,he2016deep} found that the performance of image classification can be significantly improved by using a larger number of convolutional layers and sophisticated designs. It is therefore very interesting to try and make use of similar CNN structures in speech recognition. Our previous work in \cite{Bi15} designed very deep CNNs with more convolutional layers for speech recognition, and achieved a significant reduction in WER. Subsequently  the work in \cite{Sercu16} also implemented a similar idea and verified the efficacy of very deep CNNs.

In this work, we develop a very deep CNN for acoustic modelling, and various design aspects of the architecture are explored in detail for the noisy scenario.  A comprehensive investigation and an in-depth analysis of the model are performed. Experimental results on the noisy Aurora4 task show that very promising performance can be achieved with the proposed model, even without using front-end de-noising \cite{Yu08} or sequence training \cite{vesely2013sequence}.

The rest of this paper is organised as follows. The basic CNN structure use in speech recognition is  reviewed in Section \ref{sec:basic-cnn} and the baselines are presented on the Aurora 4 task. In Section \ref{sec:verydeepcnn} a novel CNN architecture, termed the Very Deep CNN (VDCNN), is presented. Section \ref{sec:arch} explores various design aspects of the proposed VDCNN including input feature map sizes and options for padding and pooling. The use of joint adaptive training with auxiliary features is described as well as combination with a complementary LSTM-RNN acoustic model.  Finally Section \ref{sec:con} concludes the paper.

\section{Standard CNNs for ASR}
\label{sec:basic-cnn}

\subsection{Convolutional Neural Networks}
\label{scnn}
A typical CNN has two major parts: a convolutional modules followed by several fully connected layers. 
The convolutional module uses  two fundamental types of layers: the convolutional layer followed by a pooling layer.

A convolutional layer performs convolution operations to generate output values from local regions (often called receptive
fields due to the use with images) of feature maps of the previous layer, and all nodes/neurons in one
feature map share the same filter. The convolution operation can be expressed as:

\begin{equation}
    \label{eq:conv}
    \mathbf{h}^{(l)} = \sigma \left(\mathbf{W}^{(l)} * \mathbf{h}^{(l-1)} + b^{(l)} \right)
\end{equation}
where $\mathbf{h}^{(l-1)}$ and $\mathbf{h}^{(l)}$ are two feature maps in two consecutive layers. The convolutional operation (denoted as $*$) is performed with the filter $\mathbf{W}^{(l)}$ and the feature map $\mathbf{h}^{(l-1)}$. The bias $b^{(l)}$ is then added and finally the activation function $\sigma(\cdot)$, typically using a sigmoid or a rectified linear units(ReLU),  is applied to generate the the convolutional layer outputs. When multiple feature maps are present in the previous layer, all the results of convolutional operations are accumulated first before adding the bias.

The pooling layers perform down-sampling on the feature maps of the previous layer and generate new feature maps with a reduced resolution. Several pooling strategies have been investigated \cite{Sainath15a}, including max-pooling, stochastic-pooling \cite{Zeiler13}, etc, and all of them have competitive performance. In this work, max-pooling is used in all CNN models.

The most popular configuration for CNNs used in speech recognition is the setup in \cite{Sainath15a}, which has two convolutional layers with 256 feature maps in each, and it uses $9 \times 9$ filters with $1 \times 3$ pooling in the first convolutional layer, $3 \times 4$ filters in the second convolutional layer without pooling. Finally, there are  four  fully connected layers each of 2048 hidden nodes in standard multi-layer perceptron (MLP) structure. This setup is also used as the baseline CNN in this paper.
  
\subsection{Experimental setup and baselines}
To understand the behaviour of the CNNs for noise robust speech recognition, several different models are implemented and compared on the standard Aurora 4 task, which has multiple additive noise conditions as well as  channel mismatch.

The Aurora 4 task is a medium vocabulary task speech recognition task based on the Wall Street Journal (WSJ0) corpus \cite{Aurora4}. It contains 16 kHz speech data in the presence of additive noises and linear convolutional channel distortions, which were introduced synthetically to clean speech  from  WSJ0. The multi-condition training set with 7138 utterances from 83 speakers includes a combination of clean speech and speech corrupted by one of six different noises at 10-20 dB SNR, and some data is from the primary Sennheiser microphone and some are from the secondary microphone. As for the training data, the test data is generated using the same types of noise and microphones, and these can be grouped into 4 subsets: clean, noisy, clean with channel distortion, and noisy with channel distortion, which will be referred to as A, B, C, and D, respectively.

Gaussian mixture model based hidden Markov models (GMM-HMMs) are first built with Kaldi \cite{Kaldi} using the standard recipe, and consists of 3K clustered states trained using maximum likelihood estimation with the standard Kaldi MFCC-LDA-MLLT-FMLLR features. After the GMM-HMM training, a forced-alignment is performed to get the state level labels, and in the experiments here we use the synchronised clean-condition training set to generate the alignments.

The DNN/CNN hybrid baseline systems are built using CNTK \cite{CNTK}. These use 40-dimensional FBANK features with $\Delta$/$\Delta\Delta$ and an 11-frame context window. The baseline DNN consists of 6 hidden layers and each hidden layer has 2048 sigmoid nodes. The baseline CNN uses the standard CNN configuration in \cite{Sainath15a} which is also described in \ref{scnn}. All the neural networks  in this work were trained using the frame-based cross-entropy criterion (CE) objective function with stochastic gradient descent (SGD) based backpropagation (BP) algorithm. The task-standard WSJ0 bigram language model is used for decoding, and the standard testing pipelines in the Kaldi recipes are used for decoding and scoring.

The baselines results are shown in Table \ref{tab:base}. The CNN baseline is consistent with the previous work in \cite{rennie2014deep}. It is observed that CNN can get an improved performance compared to DNN, especially on the noisy subsets B, C and D, which shows the advantage of CNN in the noisy scenario.

\begin{table}[h]
\centering
\begin{tabular}{|c || c | c | c | c || c|}
\hline
Model                   & A & B & C & D & AVG \\
\hline\hline
DNN                     & 4.17 & 7.46 & 7.19 & 16.57 & 11.11  \\
CNN                     & 4.11 & 7.00 & 6.33 & 16.09 & 10.64   \\
\hline
\end{tabular}
\caption{WER (\%) of baseline DNN and CNN on Aurora 4.}
\label{tab:base}
\vspace*{-0.5cm}
\end{table}

\section{Very Deep CNNs}
\label{sec:verydeepcnn}

\begin{figure*}[t]
    \centering
    \includegraphics[width=1.0\textwidth]{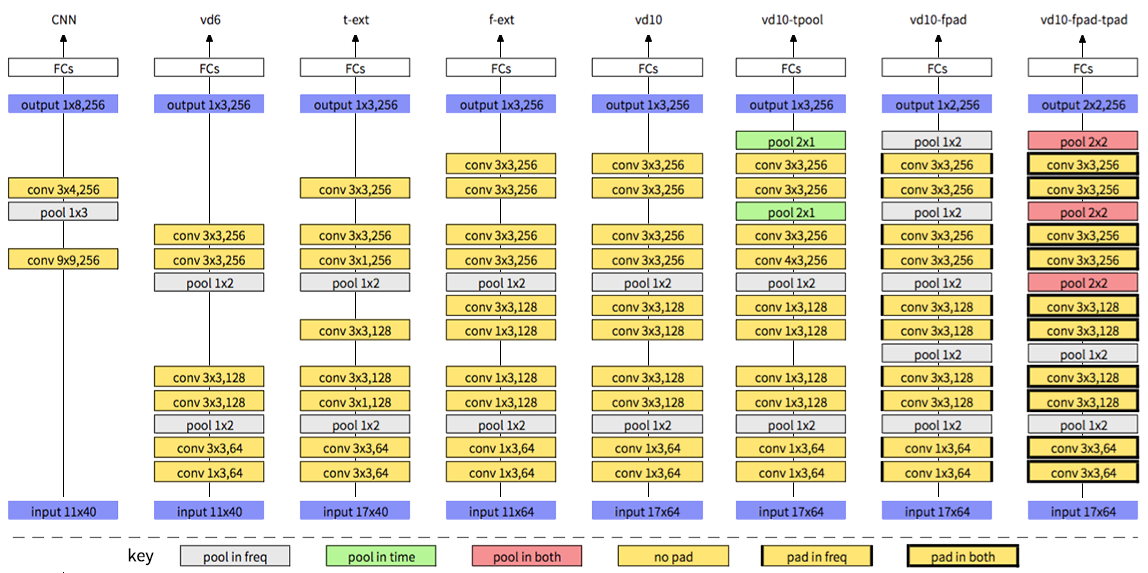}
    \caption{Very Deep CNNs Architecture}
    \label{fig:arch}
    \vspace{-0.2cm}
\end{figure*}

Our previous work in \cite{Bi15} introduced very deep CNNs for ASR of conversational telephone speech, and verified the promising potential of this kind of  model. Based on this initial work, the models are here further developed with a focus on improving performance for noisy speech. Before the detailed architecture is described, some fundamental principles of VDCNNs for speech recognition are presented and explained:

\begin{itemize}

\item Rather than using the traditional $9 \times 9$ or $3 \times 4$ filters and $1 \times 3$ pooling \cite{Sainath13,sainath2015convolutional,Sainath15a}, VDCNNs for speech recognition use filters of $3 \times 3$ (sometimes $1 \times 3$ and $3 \times 1$), and the pooling size is constrained to $1 \times 2$ or $2 \times 2$. This is also similar as the work in image classification\cite{Simonyan14}. The stride of the convolution is set to $1$ and only non-overlapping pooling is used in this work. These are the smallest reasonable sizes for the filters and pooling respectively \cite{Simonyan14}, which enables the models to be built with as many convolutional layers as possible .

\item Compared to computer vision tasks, the size of neural network inputs in speech recognition is relatively small since the  context window and basic feature (e.g. FBANK) dimension are much smaller in most of speech recognition systems\footnote{Normally $11 \times 40$ inputs are used, 40-dimensional FBANK features with 11 consecutive frames.}. Accordingly, in addition to the adjustment of the size of filters and pooling, the size of inputs need to be enlarged appropriately for speech recognition to allow more convolution and pooling operations. All the proposed very deep CNNs in this work only  one input feature map, i.e. the static feature map,  is used unless otherwise noted.

\item For the very deep CNNs, a pooling layer is added after at least two convolutional layers. The feature map size before the first fully-connected layer is set to a relatively small value in our proposed VDCNNs, e.g. $1 \times 3$ or $1 \times 2$. In addition the number of feature maps is increased gradually and doubled after some pooling layers. For all the model configurations in this paper, the number of feature maps is increased with subsequent layers and in sequence uses  64, 128 and then 256 feature maps.

\item As well  convolutional layers and associated pooling layers, 4 fully-connected (FC) layers are added, with an  output layer using a softmax activation function. The ReLU is used for the hidden nodes in all proposed VDCNNs to overcome the gradient-vanishing problem.

\end{itemize}

Following these fundamental principles, the very deep CNNs are developed. One model, named {\tt vd6}, is introduced as the first very deep CNN model, whose configuration can be found in Figure~\ref{fig:arch}. The model {\tt vd6} shares the same context window size and feature dimension, i.e. $11 \times 40$, with the traditional CNN shown as the first column of Figure~\ref{fig:arch}. Following these fundamental principles, 5 convolutions can be performed in time, and 6 convolutions and 2 poolings can be performed in frequency, which results in a VDCNN with 6 convolutional layers and 2 pooling layers. Further modifications are based on {\tt vd6}, and all the discussed structures in this work are illustrated in Figure~\ref{fig:arch}: different types of pooling layers are indicated with different colours, and convolutional layers with various padding strategies are marked with corresponding border styles.

\section{Architecture}
\label{sec:arch}

In  this  section,  detailed  investigations of  the  VDCNN architecture will be performed for robust speech recognition.

\subsection{Input feature map extension}
The typical size of input feature maps in speech recognition is $11 \times 40$, which is much smaller than that used in images. Accordingly, the context window and frequency dimension extension in the input feature maps are firstly explored. Following {\tt vd6} with 5 convolutions in time and 6 convolutions in frequency, {\tt t-ext} and {\tt f-ext} are developed with extended input feature maps. In model {\tt t-ext}, the context window size is extended to 17 frames allowing 8 convolutional layers stacked in time with a filter size of $3$, and {\tt f-ext} enlarges the FBANK to 64 dimensions which allowing 10 convolutional layers to be stacked. As shown in Table~\ref{tab:extension}, the first line is the traditional baseline CNN and the other lines are the proposed very deep CNNs. Increasing the number of convolutional layers can obtain a large reduction in WER, and both {\tt t-ext} and {\tt f-ext}, in which the extended input feature maps can contribute more convolutional layers, get further significant improvement over the model {\tt vd6}.

Finally the input extension is performed in both time and frequency with $17 \times 64$, and the model is named {\tt vd10} to indicate that it has 10 convolutional layers. {\tt vd10} in Table~\ref{tab:extension} shows that the benefits from both {\tt t-ext} and {\tt f-ext} can be combined. Moreover most of the contributions are from the three noisy subsets B, C and D, which demonstrates the effect of the proposed VDCNNs for noise robust speech recognition.


\begin{table}[h] \small
\centering
\begin{tabular}{|c | c | c || c | c | c | c || c|}
\hline
Model                   & T $\times$ F & L & A & B & C & D & AVG \\
\hline\hline
CNN                     & $11 \times 40$ & 2 & 4.11 & 7.00 & 6.33 & 16.09 & 10.64   \\
\hline
\tt vd6                 & $11 \times 40$ & 6 & 3.94 & 6.86 & 6.33 & 15.56 & 10.34   \\
\tt t-ext            & $17 \times 40$ & 8 & 3.72 & 6.57 & 5.83 & 14.79 & 9.84   \\
\tt f-ext            & $11 \times 64$ & 10 & 3.79 & 6.51 & 6.26 & 15.19 & 10.02   \\
\tt vd10                & $17 \times 64$ & 10 & 4.13 & 6.62 & 5.92 & 14.53 & \bf 9.78   \\
\hline
\end{tabular}
\caption{WER (\%) comparisons of the models with the context window and frequency dimension extension. {\bf F} indicates the size on {\bf F}requency axis and {\bf T} indicates the size on {\bf T}ime axis. {\bf L} indicates the number of convolutional layers}
\label{tab:extension}
\end{table}

\begin{table*}
\centering
\begin{tabular}{|l | c | c c || c | c | c | c || c|}
\hline
Model                   & \#    & Pooling & Padding & A & B & C & D & AVG  \\
\hline\hline
\tt vd10                & \multirow{4}{*}{1}  &  F  &  ---    & 4.13 & 6.62 & 5.92 & 14.53 & 9.78   \\
\tt vd10-tpool          &   & F \& T   &  ---    & 3.68 & 6.46 & 6.13 & 15.03 & 9.91   \\ 
\tt vd10-fpad           &  &  F  &  F    & 3.57 & 6.17 & 5.31 & 14.24 & 9.38   \\
\tt vd10-fpad-tpad      &   &  F \& T  &  F \& T    & 3.27 & 5.61 & 5.32 & 13.52 & \bf 8.81   \\
\hline
\tt vd10-fpad-tpad      &   3 &  F \& T  &  F \& T    & 3.79 & 6.11 & 5.60 & 13.62 & 9.13   \\
\hline
\end{tabular}
\caption{WER (\%) comparison of the proposed very deep CNNs with different architectures. {\bf F} indicates the {\bf F}requency axis and {\bf T} indicates the {\bf T}ime axis. {\bf \#} indicates the number of input feature maps}
\label{tab:pooling}
\end{table*}

\subsection{Pooling in time}

All of these previous VDCNNs use pooling in frequency and do not use pooling in time. As stated in \cite{Sainath13}, pooling in time may result a degradation in the standard shallow CNN system. To investigate whether pooling in time for very deep CNNs with a wider input and a deeper structure is useful, more experiments are conducted. The {\tt vd10-tpool} is a model with 2 temporal pooling stages. From Figure~\ref{fig:arch} we can see, compared to {\tt vd10}, {\tt vd10-tpool}  two more temporal pooling stages are added (temporal convolutions are added  to keep the sizes of input and output feature maps unchanged). The first two lines in Table~\ref{tab:pooling} show that pooling in time does not help the VDCNN {\tt vd10}, which is a similar conclusion to the shallow CNN \cite{Sainath13}.

\subsection{Padding in feature maps}

In most work on CNNs for speech recognition, the convolutions are performed without padding, including our previous work \cite{Bi15} on the very deep CNNs and the work on traditional CNNs \cite{Sainath13,Sainath15a,Sainath15b}. However for computer vision, including the work \cite{Szegedy14,Simonyan14}, convolutions are usually performed after zero-padding the feature maps. It can save the size of feature maps so that it is a useful way to increase the model depth. Padding in feature maps can better use the border information of feature maps by the neural network, which is beneficial for the final performance \cite{Yoshioka16}.

In this work, we tried to use padding in the very deep CNNs for speech recognition. Based on the model {\tt vd10}, the VDCNNs with different padding strategies were implemented. {\tt vd10-fpad} indicates the model with padding only in frequency, and as a result it can perform more pooling operations in frequency compared to {\tt vd10}. In addition, padding in both dimensions is also applied, indicated as {\tt vd10-fpad-tpad}. In this model, considering that pooling is a necessary approach to reduce the feature map size to a reasonable value when doing padding in time, so pooling in time is also applied. The model {\tt vd10-fpad} and {\tt vd10-fpad-tpad} are illustrated as the last two columns of Figure~\ref{fig:arch}.

The middle two lines of Table~\ref{tab:pooling} shows the results with different padding strategies. It shows that {\tt vd10-fpad} is significantly better than {\tt vd10} by padding in frequency, and {\tt vd10-fpad-tpad} padding in both time and frequency obtains a further improvement. These results give the conclusion that padding in feature maps for the very deep CNNs is very important to better encode the border information, which results in a lower WER. Compared to the traditional CNN, we can see that {\tt vd10-fpad-tpad} containing up to 10 convolutinal layers with appropriate padding and pooling strategies achieves a very large reduction in WER for noise robust speech recognition.

\subsection{Input feature maps selection}

All the proposed very deep CNNs are using one feature map as input, i.e. the static FBANK features. Considering that most of the published works used three feature maps for speech recognition (including the dynamic features, $\Delta$ and $\Delta\Delta$), the number of input feature maps are compared for the very deep CNN in the noisy scenario, and the related results within the very deep CNN are shown as the last line of Table~\ref{tab:pooling}.

It is interesting to find that the one input feature map based VDCNNs are clearly  better than the model using three input feature maps for noisy scenarios, and this conclusion is different from the previous one from the shallow CNN models \cite{Sainath13}. One possible explanation would be that the dynamic feature may have less information than the static one, and the knowledge can be better extracted from the raw static features directly by the very deep CNNs. In addition, some information, existing in dynamic features, can also be well captured using very deep CNNs with a wider context window.

\subsection{Joint training of VDCNNs with auxiliary features}
\label{sec:adap}

The use of auxiliary features in factor-aware training is one type adaptation popular for robust ASR \cite{tan2016speaker,seltzer2013investigation,multi-factor-Qian2016,qian-tasl16}. Considering that topographical features, such as FBANK, are more appropriate than the non-topographical feature, such as fMLLR and i-vector, for the CNN usage \cite{soltau2014joint}, an auxiliary feature joint training framework is proposed within the very deep CNN architecture, shown as the left part of Figure \ref{fig:jtjd}. Rather than the normal approach that concates the auxiliary features with the FBANK to be fed into the neural networks \cite{tan2016speaker}, here the auxiliary features are separately transformed with a normal fully-connected layer first, which is parallel to the FBANK-based VDCNN block, and then the hidden layer outputs of these two parallel streams are concatenated into the following shared fully-connected hidden layers.

For the auxiliary features, the use of MFCCs, fMLLR features and i-vector are explored in this work. The normal 39-dim MFCC is used with static, $\Delta$ and $\Delta\Delta$ features, and 40-dim fMLLR features are generated using all the data from a given speaker. An 11-frame context window is used for MFCC and fMLLR. For the i-vector, a GMM with 2048 Gaussians is used to extract a 100-dimensional i-vector for each utterance, and these i-vectors are obtained using fMLLR features. The results of joint training with various auxiliary features are shown in Table \ref{tab:adap}. The appropriate auxiliary feature type is important: joint learning using MFCC gets no more improvement, in contrast the adapted auxiliary features can achieve another large gain based on the advanced very deep CNN. Most of the improvements are from the three noisy subsets, so this shows that the adaptation is very important for noise robust speech recognition.

\begin{table}[h] \small
\centering
\begin{tabular}{|c | c || c | c | c | c || c|}
\hline
Model                   & Aux & A & B & C & D & AVG \\
\hline\hline
\multirow{4}{*}{\footnotesize{VDCNN}}                     & --- & 3.27 & 5.61 & 5.32 & 13.52 & 8.81   \\
 & \footnotesize{MFCC} & 3.14 & 5.66 & 5.27 & 13.55 & 8.83\\
 &  \footnotesize{fMLLR} & 3.03 & 5.46 & 4.45 & 12.22 & 8.11 \\
 & \footnotesize{fMLLR} + ivec & 3.08  & 5.27 & 4.54 & 12.11 & \bf 7.99 \\
\hline
\end{tabular}
\caption{\small{WER (\%) comparisons of the very deep CNNs joint learning with various auxiliary features. {\bf VDCNN} indicates the previous best very deep CNN, and {\bf Aux} indicates the auxiliary feature types}}
\label{tab:adap}
\end{table}


\subsection{Joint decoding of VDCNN and RNN}
\label{sec:jd}

Recently the use of long short-term memory recurrent neural networks (LSTM-RNN) based acoustic model has shown great promise on several tasks \cite{Sainath15b,sak2014long,sak2014sequence}. Considering the very deep CNN is one type of feed-forward neural networks,  it is interesting to investigate its complementarity with  an LSTM-RNN. Hence, an LSTM-RNN was also built and it uses a single frame of 40-dimensional FBANK features with 5 frames shift as input, and the model has three LSTMP (LSTM-Projected) \cite{sak2014long} layers, where each LSTMP layer has 1024 memory cells and 512 hidden nodes in the projection. Truncated back propagation through time (BPTT) is used to train the LSTM model with the chunk size set to 20 frames, and 40 utterances are processed in parallel to form a mini-batch. To ensure the stability of training, the gradient is clipped to the range of $[-1,1]$ during parameter update. In addition, a speaker-aware LSTM-RNN using 100-dimensional i-vector is also constructed in a similar way to  the work in \cite{tan2016speaker} using concatenated features as the inputs. The results at the top part of Table \ref{tab:jd} show that LSTM-RNN can get similar performance as the baseline shallow CNN on this task, and speaker-aware training using i-vectors achieves a significant reduction in WER.

\begin{figure}[t]
    \centering
    \begin{tabular}{ccc}
    \subfloat[Joint training of VDCNNs with auxiliary features]{
        \label{fig:jt}
        \includegraphics[height=5.5cm,width=0.21\textwidth]{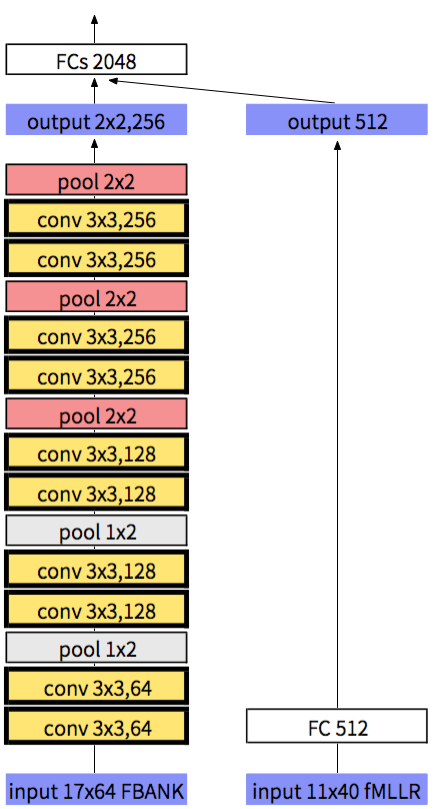}
    } &
    \subfloat[Joint decoding of VDCNN \& RNN]{
        \label{fig:jd}
        \includegraphics[height=5.5cm,width=0.21\textwidth]{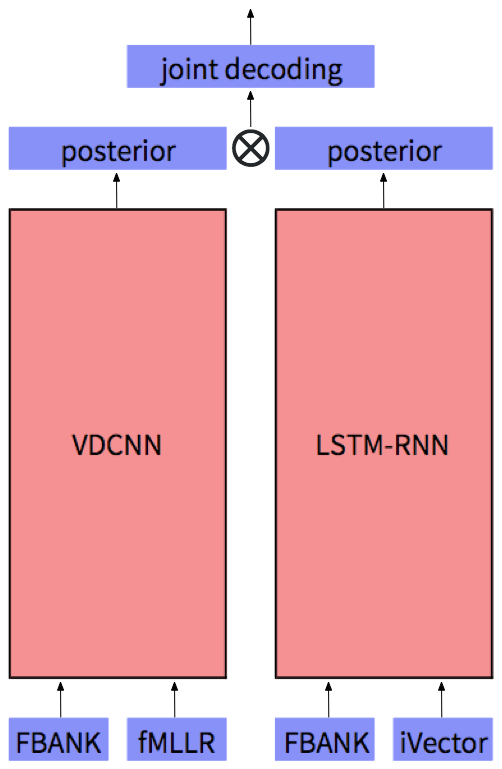}
    }
    \end{tabular}
    \caption{The architectures of VDCNN with auxiliary features joint training, and VDCNN \& RNN joint decoding}
    \label{fig:jtjd}
\end{figure}

As one way to combine the two acoustic models, we implemented a joint decoding scheme shown as the right part of Figure \ref{fig:jtjd}, which uses a weighted combination of state-level acoustic log likelihoods from the VDCNN and LSTM-RNN systems. Note that we used the same decision tree for these two systems as the work in \cite{wang2015joint,woodland2015cambridge}, and the symbol $\otimes$ denotes joint decoding. In this work, the adapted VDCNN and adapted LSTM-RNN systems, shown as \textcircled{\footnotesize{1}} and \textcircled{\footnotesize{2}} in the top part of Table \ref{tab:jd}, are used for joint decoding. Considering that the performance of VDCNN is better, a slightly biased weight pair, 0.6 and 0.4 for VDCNN and RNN respectively, is used here. For comparison, the normal Kaldi minimum Bayes risk (MBR) lattice combination is also applied with these two systems and denoted with symbol $\oplus$. The results of both these two approaches are shown in the bottom part of Table \ref{tab:jd}.

It is interesting to observe that although the performance gap within these two systems is very large (almost 2.0\% absolute), the combination still can obtain a very promising improvement, which demonstrates the huge complementarity within VDCNN and RNN. Comparing the different combination methods, the proposed joint decoding scheme obviously achieves a better performance than the MBR lattice combination.

\begin{table}[h] \small
\centering
\begin{tabular}{|c | c || c | c | c | c || c|}
\hline
Model                   & Aux & A & B & C & D & AVG \\
\hline\hline
\textcircled{\footnotesize{1}} \footnotesize{VDCNN}                 & fmllr/ivec & 3.08  & 5.27 & 4.54 & 12.11 & 7.99   \\
\multirow{2}{*}{\footnotesize{LSTM-RNN}}  & --- & 3.92 & 7.21 & 6.63 & 15.94 & 10.68  \\
& \textcircled{\footnotesize{2}} ivec & 3.64 & 6.81 & 6.28 & 14.49 & 9.84  \\
\hline
\textcircled{\footnotesize{1}} $\oplus$ \textcircled{\footnotesize{2}} & \multirow{2}{*}{fmllr/ivec}  & 2.86 & 4.82 & 4.26 & 11.27 & 7.41 \\
\textcircled{\footnotesize{1}} $\otimes$ \textcircled{\footnotesize{2}} &  & 2.82 & 4.67 & 4.17 & 10.70 & \bf 7.09 \\
\hline
\end{tabular}
\caption{\small{WER (\%) comparisons of VDCNN/RNN joint decoding. {\bf $\oplus$} indicates the normal MBR lattice combination using Kaldi, and {\bf $\otimes$} indicates the state-level weighted log likelihood score combination in  joint decoding.}}
\label{tab:jd}
\vspace*{-0.5cm}
\end{table}

\subsection{Evaluation summary on Aurora 4}
\label{exp:aurora4}

Finally, in Table \ref{tab:sum}, the results obtained using the proposed new models are summarised. It is observed that compared to the standard shallow CNN baseline, the performance can be greatly improved by increasing the convolutional layer depth (17.0\% relative WER reduction), and gets significant gains on all subsets. Furthermore based on the proposed very deep CNN, adaptation using the auxiliary feature assisted joint training, and joint decoding with an LSTM-RNN model can obtain further very large improvements on this noisy data, giving  25.0\% and 33.0\% relative WER reductions  compared to the baseline.

\begin{table}[h] \small
\centering
\begin{tabular}{|c || c | c | c | c || c|}
\hline
Model                   & A & B & C & D & AVG \\
\hline\hline
CNN                     & 4.11 & 7.00 & 6.33 & 16.09 & 10.64   \\
\hline
VDCNN                 & 3.27 & 5.61 & 5.32 & 13.52 & 8.81   \\
+ Aux                & 3.08  & 5.27 & 4.54 & 12.11 & 7.99 \\
$\otimes$ LSTM-RNN  & 2.82 & 4.67 & 4.17 & 10.70 & \bf 7.09 \\
\hline\hline
AD OSN LRF \cite{rennie2014deep} & 4.0 & 7.2 & 6.4 & 14.5 & 10.0 \\
AD CNN Adpt \cite{swietojanski2016learning} & 3.4 & 5.7 & 6.1 & 13.4 & 8.7 \\
\hline
\end{tabular}
\caption{\small{WER (\%) comparison of various systems on Aurora 4}}
\label{tab:sum}
\end{table}

The results are also compared with some other published systems on Aurora 4 in the literature shown as the bottom of Table \ref{tab:sum}. One is without adaptation and the other uses adaptation (adapted on both speaker (83 speakers) and environment (6 noisy scenarios) levels). The systems developed here  outperform the others both with and without adaptation.

\section{Conclusion}
\label{sec:con}

This paper has extended our previous work on  very deep CNNs for noise robust speech recognition. Compared to the standard shallow CNN structure normally in speech recognition, the sizes of filters and pooling operations are constrained to be small and the input feature maps are made larger. This adjustment enables us to have up to ten  convolutional layers. A detailed analysis on the appropriate setup for pooling, padding and input feature map selection have been performed. The results show that time pooling without padding is not useful, but padding on both time and frequency dimensions of feature maps is effective for very deep CNNs. Compared  to  the  traditional  input  feature map  usage  with  dynamic  features,  very  deep  CNNs  only using  the  static  features  are  more  effective  for  noise  robust speech  recognition. In addition, an adaptation framework for very deep CNNs joint training with auxiliary features is proposed to get further reductions in WER.  Further improvements from building a complementary LSTM-RNN model is demonstrated. In particular it is shown that using a state-level weighted log likelihood score combination in a joint acoustic model decoding scheme is more effective than  the standard Kaldi MBR lattice combination.

The proposed new model is evaluated on the Aurora4 with additive noise and channel mismatch. The single best very deep CNN system achieves a WER of 8.81\%, and further 7.99\% with auxiliary feature joint training. The final joint decoding scheme with LSTM-RNN gives a  WER of 7.09\%. At the time of writing, to our knowledge this is the best published results on the Aurora 4 task, even without feature enhancement or sequence training.

\bibliographystyle{IEEEbib}
\bibliography{strings,refs}

\end{document}